# Hybrid approach to detecting symptoms of depression in social media entries.

*Submission Type: Completed Research*

**Agnieszka Wołk**
[1]Polish-Japanese Academy of Information Technology
[2]The Institute of Literary Research of the Polish Academy of Sciences
[1]Koszykowa 86, 02-008 Warsaw
[2]Nowy Świat 72, 00-330 Warsaw
awolk@pjwstk.edu.pl

**Karol Chlasta**
Kozminski University
Jagiellońska 57/59, 03-301 Warsaw
kchlasta@kozminski.edu.pl

**Paweł Holas**
University of Warsaw
Krakowskie Przedmieście 26/28, 00-927 Warsaw
pawel.holas@psych.uw.edu.pl

**Abstract**

*Sentiment and lexical analyses are widely used to detect depression or anxiety disorders. It has been documented that there are significant differences in the language used by a person with emotional disorders in comparison to a healthy individual. Still, the effectiveness of these lexical approaches could be improved further because the current analysis focuses on what the social media entries are about, and not how they are written. In this study, we focus on aspects in which these short texts are similar to each other, and how they were created. We present an innovative approach to the depression screening problem by applying Collgram analysis, which is a known effective method of obtaining linguistic information from texts. We compare these results with sentiment analysis based on the BERT architecture. Finally, we create a hybrid model achieving a diagnostic accuracy of 71%.*

## Introduction

Depression is not only a most prevalent psychiatric disorder, but also a recurrent and burdensome condition and one of the leading causes of disability worldwide (World Health Organization, 2008).

Diagnosis of affective disorders, to which category depression belongs, requires filling standardized questionnaires and the conducting of structured clinical interviews with patients by mental health specialists. The process usually spreads in time over weeks (Beck et al. 1993) We believe that the diagnosis process can be supported by application of methods based on artificial neural networks. For example, (McGinnis et al. 2018) proposed the use of a 90-second fear induction task during which participants' motion is monitored using a wearable sensor. Machine learning and data extracted from the most clinically feasible 20-second phase of the task were used to predict depression and anxiety disorder in a sample of children with and without an internalizing diagnosis. The best performing logistic regression provided a diagnostic accuracy of 80%.

Early diagnosis of attentional biases underlying affective disorders could help to target people at risk for developing them. Once diagnosed, these biases could be reduced by means of Attention Bias Modification Training (ABMT) or mindfulness training (Holas et al. 2020). ABMT was found to change the attentional patterns of people who are depressed (Krejtz et al. 2018), and to reduce anxiety vulnerability and ameliorating dysfunctional anxiety (MacLeod et al. 2012). It is doing so by redirecting attention and promoting positive attention bias among vulnerable individuals, such as clinically depressed (Krejtz et al. 2018).

To date, natural language processing (NLP) researchers have proven that machine learning can be used to detect symptoms associated with depression by conducting lexical and predictive analyses of text created by people on the Internet, especially in social media entries (Wolohan et al. 2018). Studies showed significant differences in language used by individuals suffering from depression, as well as differences in the ability of machine learning algorithms to correctly detect symptoms of depression in topic-restricted texts (Islam et al. 2018; Jung et al. 2017; Zucco et al. 2017). As authors explain, a limited efficacity might be caused by the ability of writers to avoid discussing depression to hide their potential mental health (e.g. to avoid self-stigmatization).

There are several reasons why social media entries might not be classified correctly in an automated way. The texts might describe depression in general, or the writer might purposefully control what is shared in social media. The same applies, for example, to nostalgic texts and book reviews on certain subjects that appear to be 'depressive' or 'unhappy'. The correct classification of sentiment and emotions in social media entries is still a challenge of NLP (Al Hanai et al. 2018; Morales et al. 2016). Therefore, we believe that new approaches to automated analysis of text are needed to improve computer-based assessment of depression through increased natural language understanding by computers.

In the current study, we apply a novel approach in which we focus on how texts are constructed and how they were created by applying Collgram analysis. We departed from a typical approach to sentiment analysis (SVM—support vector machines) and refocused on a more innovative approach based on the BERT architecture (Wołk 2019), which, according to the literature, gives more accurate results. We treat the SVM method as a baseline in our study, upon which we want to improve our classification results by applying the BERT architecture (Sun et al. 2019). To the best of our knowledge, this approach has not yet been previously used in relation to the study of depression.

BERT (Bidirectional Encoder Representations from Transformers) is an NLP model developed by Google for pre-training language representations. It leverages an enormous amount of publicly available plain text data on the web (e.g. from Wikipedia and Google Books) and is trained in an unsupervised manner. It is a powerful model that is trained to learn a language's structure and its nuances by training a language model. BERT (Devlin et al. 2019) has a deep and bi-directional structure to it unlike ELMo (Peng et al. 2019), which is shallow and bi-directional, while OpenAI GPT (Lee et al. 2020) is uni-directional in nature. This bi-directional nature helps the model to capture the context from both previous and subsequent words at any given time.

Collgram profiles can be a complementary technique for text analysis (Bestgen et al. 2018). They were originally used widely to study the phraseological competences of people who speak a foreign language, although it is increasingly being used in longitudinal research and machine learning evaluation. By analysing the assumptions of this technique, we concluded that it can be used to detect symptoms of anxiety, depression or suicidal tendencies based on social media entries.

We believe that the classification result obtained by our hybrid model based on BERT and supported by Collgram profiles as the next layer of analysis, will improve classification results for detecting symptoms of depression in social media entries (containing between 150 and 250 words).

# Methods

Polish language presents a significant challenge in NLP (Wołk et al. 2017). We re-implemented and retrained the original BERT model, and built the Collgram profile creation tool. In addition, we have also enriched Collgram profiles with additional measures potentially increasing the precision for complex languages—in particular, DICE and Lexical Gravity (Daudavicius et al. 2004).

To conduct the experiments, we trained our model on the data from the Common Crawl (Bevendorff et al. 2018) project. As the CommonCrawl corpus is very polluted, an algorithmic division into sentences, data de-duplication, and cleaning was necessary.

The original corpus was 296 GB in size and consisted of 1,962,047,863 sentences. After the data pre-processing its size was reduced to 94 GB and 920,517,413 sentences. The details about the exact number of n-grams used in our Collgram analysis are provided in Table 1.

| Order | n-grams |
|---|---|
| 1-grams | 421,344,934 |
| 2-grams | 2,978,853,249 |
| 3-grams | 2,003,857,026 |
| 4-grams | 2,270,689,390 |
| 5-grams | 2,088,765,597 |
| 6-grams | 1,844,154,756 |

**Table 1. n-grams amount in corpus.**

### *BERT Model*

We trained the BERT model on all the data available in the CommonCrawl project. We fine-tuned and retrained the trained model. Since we did not have a sufficiently large corpus of texts strictly associated with depression, we prepared our own dataset using a web crawler. This dataset was based on the opinions of users assigning ratings to online stores (e.g. euro.com.pl), providing opinions about doctors (znanylekarz.pl) evaluating work at companies (gowork.pl) and evaluating online stores and their products (opineo.pl). In this way, 200,000 different opinions were obtained, out of which 10,000 were randomly selected for testing. The dataset was balanced while crawling in such a way that it contained the same number of positive and negative opinions. The whole process was performed based on https://github.com/google-research/bert/blob/master/predicting_movie_reviews_with_bert_on_tf_hub.ipynb (Devlin et al. 2019). The precision of the created model on the test sample was 0.91, and the F1 score was 0.89.

### *Collgram analysis*

The technique implemented in this paper assigns each n-gram (i.e. any adjacent pair of words in L2 texts) the results of the association calculated based on the reference corpus. The resulting units are referred to as CollGrams. CollGram was proposed by Yves Bestgen and Sylvianne Granger in 2014 (Bestgen et al. 2014). It produces three measures—a mutual information factor (MI), t-score and the number of idiosyncratic units—that together form the CollGram profile, according to the authors determines the collocation strength of each text.

The next step in our method is to calculate these three measures to estimate the collocation strength of each text automatically:

- The mean MI result, which measures collocations consisting of rare words.
- The average t-score, which measures collocations consisting of very common words.
- Proportions of bi-grams, which are not present in the reference corpus, and therefore cannot be attributed to any association result. These bi-grams can be mistakes or creative combinations.

The mutual information (MI) factor (Song et al. 2012) has come to gradually occupy a central place in corpora-based linguistics, because it is one of the first methods to measure the possibility of combining words and is easy to apply.

The MI for each pair is calculated according to the following formula:

$$MI(x,y) = \log_2 \left( \frac{N \cdot f(x,y)}{f(x) \cdot f(y)} \right)$$

where f(x,y) is the frequency of coexistence of x and y, f(x) and f(y) are the respective frequencies of x and y in any place in the text, and N is the corpus size. If x and y tend to occur in combination, their mutual information will be high. If they are not related and appear only by coincidence, their mutual information will be low. In other words: the greater the mutual information, the stronger the relationship between two words.

The smaller the sum of the frequencies, the greater the MI, and vice versa. This correlation and analysis of the MI pattern makes it possible to conclude that MI enhances those pairs that are made up of words with similar frequencies by placing them at the top or bottom of the frequency list. Two coexisting rare words will indicate a high score, but two common words will have a low score.

The T-score (Lenko-Szymanska et al. 2016) presents a different way of calculating the combination of word pairs based on the frequency of each word in combination:

$$T(x,y) = \frac{f(x,y) - \frac{f(x) \cdot f(y)}{N}}{\sqrt{f(x,y)}}$$

where f(x,y) is the frequency of words in pairs, f(x) is the frequency of the first word, f(y) is the frequency of the second word, while N is the total size of the input dataset. The T-score shows the following correlations between the combination and the sum of the word frequencies for word pairs: the smaller the sum of the frequencies, the smaller the connection range and T-score, and vice versa—the greater the sum of the frequencies, the greater the range and T-score. The range of possibilities for combining from minimum to maximum is determined by the frequency of the word pair, because the frequencies of these word pairs, which are often used together, play the most important role in the pattern. Therefore, the T-score highlights frequent pairs of words that have high frequency sums. For the same frequency sums, the T-score increases when the corresponding word frequency in the pair is significantly different and decreases when they are similar. Therefore, only these words that form pairs with individual high frequencies register high on the curve.

The average, minimum, and maximum levels of the T-score increase as the frequency of word pairs increases, but not as significantly as in the case of MI.

The T-score is more flexible than MI because it is possible to isolate collocations that have a broader range of ability to create collocations, i.e. they are used less and less often. MI includes a narrower scale of the sum of the word frequencies and thus, fewer word combinations than the T-score.

As part of this work, the Collgram profile has been extended by two additional measures. The first is the DICE measure (De schuymer et al. 2003), which can be used to calculate the co-occurrence of words or a group of words. It is defined according to the following formula:

$$Dice(x,y) = \frac{2 \cdot f(x,y)}{f(x) + f(y)},$$

where f(x,y) is the frequency of co-occurrence of x and y, and f(x) and f(y) are the respective frequencies of x and y in any place in the text. If x and y tend to occur in combination, their DICE result will be high. To observe many small numbers, a logarithm is added to this formula. Thus, the formula is slightly modified to better observe the correlation between the sum of the frequencies and the DICE result. The ability to combine each pair of words using this calculation method is measured by the following formula:

$$Dice(x,y) = log_2 \left( \frac{2 \cdot f(x,y)}{f(x) + f(y)} \right)$$

Two coexisting rare words will always record a high score, but two common words may register a low or high score. Unlike the MI result, the highest DICE value is similar over the entire frequency sum range. This is because the corpus size is not represented in the formula. The correlation shows that the DICE and MI results are very similar, and the size of the corpus in the formula does not play a significant role, which is extremely important for morphologically rich languages.

All methods used to calculate the ability to combine word pairs discussed above largely depend on the frequencies of the component words. However, significant collocation can consist of both frequent and rare words, because the relationship between the words is an important factor. Regular collocations between objects that occur more often than their respective frequencies, and the length of the text in which they occur, might be predictable. It is important to consider the mutual expectation of words, not only their frequency. Therefore, a new method of collocation extraction has been introduced, which is less dependent on the frequency of words. This method is called Gravity Counts (Wolohan et al. 2018). In this approach, the extent of collocation creation is the same across the scale of possible frequency sums. In other words, the average, maximum and minimum levels are the same for all frequency sums to identify the boundary at which word combinations change into collocations.

The Gravity Counts method is based on the assessment of the possibility of combining two words in a text that takes into account various frequency features, such as the frequency of individual words, the frequency of word pairs, and the number of different words in the selected range. The last feature is new i.e. it does not appear in the analysed statistical tools. The Gravity Counts method emphasizes the co-occurrence of two words in the text within a selected range—in this case, three words. If the first word, x, is used more often than expected before the second word y, and the second word y is used more often than expected after the first word x, then x and y form a collocation. This concept is expressed in the following form of Gravity Counts:

$$G(x,y) = log \left( \frac{f(x,y)}{g(x)} \cdot \frac{f(y)}{g'(y)} \right)$$

where f(x, y) is the frequency of the pair of words x and y in the corpus, g(x) is the word diversity factor on the right side of x, and g'(y) is the word diversity factor on the left side of y. The ratio of the word pair frequency and the diversity of the surrounding words help in assessing co-occurrence. If the frequency of a pair of words is much greater than the variety of surrounding words, then this pair of words can be treated as a collocation. The g(x) and g'(y) terms are calculated as follows:

$$g(x) = \frac{f(x)}{n(x)} \qquad g'(x) = \frac{f(y)}{n'(y)}$$

where n(x) is the number of different words to the right of x, f(x) is the frequency of x in the corpus, n(y) is the number of different words to the left of y, and f(y) is the frequency of y in the corpus. The diversity factor shows the relationship between a given word and the variety of words surrounding it.

The higher the frequency of a pair of words over ordinary diversity, the higher the value of their combination, and vice versa.

The final expression of Gravity Counts (Daudavicius et al. 2004) is as follows:

$$G(x,y) = \log\left(\frac{f(x,y) \cdot n(x)}{f(x)}\right) + \log\left(\frac{f(x,y) \cdot n'(y)}{f(y)}\right)$$

Gravity Counts can be applied to words of any frequency because the range of combination options is similar for rare and common words.

As in the case of BERT method described earlier, the same corpus—based on the CommonCrawl project—was used to train the reference model for the Polish language.

Both solutions were normalized to return an easy-to-interpret result on a scale of 0 to 100, where 0 is a completely negative and 100 is a completely positive result. The solutions have been implemented as online services and are publicly available on the internet[1], as well as in offline form on GitHub[2].

## Data preprocessing

Obtaining an adequate amount of test data to conduct the study was a significant challenge. There are sources containing the desired data, like the Polish Corpus of Suicide Farewell Letters (Zaśko-Zielińska et al. 2015), but these are not publicly available. The challenge lay also with applying self-reporting tools, and a risk that a large percentage of people with depression can be misdiagnosed. When screening with the PHQ-8 questionnaire (Kroenke et al. 2009), clinical practitioners recognize only about 65% of depressive cases accurately. To mitigate the risks and improve the quality of our data, each diagnosis was repeated three times. The texts from our corpus were gathered from social media: Facebook (78 entries), and Reddit (69 entries). These entries were generated by subjects who declared a depressive disorder, which was reconfirmed by a clinical interview. Medical cooperation was established through the Polish Telemedicine and e-Health Association with psychologists and psychiatrists from the Medical University of Gdansk. Two teams were created remotely, conducting clinical interviews with the authors of the social media entries, and making independent verification of self-reported measures. This data was then shared with authors for the purpose of this research.

It was ultimately determined that 131 entries were created by people with major depressive disorder and 16 by people who were considered healthy. The collections were balanced by adding 115 entries written by healthy people. We standardized all the text to contain between 150 and 250 words. They were tokenized and normalized before being evaluated by our models.

## Results

Using the classifier based on SVM, we were able to classify 63% of depressive social media entries correctly. 17% of texts generated by depressive subjects were misdiagnosed, and the 20% of entries generated by healthy individuals were misdiagnosed as depressive. As all our models were normalized to return results in the range of 0 to 100—where 0 indicates full depression and 100 indicates a text generated by a healthy individual—it is possible to set a limit that discriminates between healthy and unwell patients. In this case, the best result for such division of space was obtained when this value was 53, as shown in Figure 1.

---

[1] http://collgram.pja.edu.pl

[2]

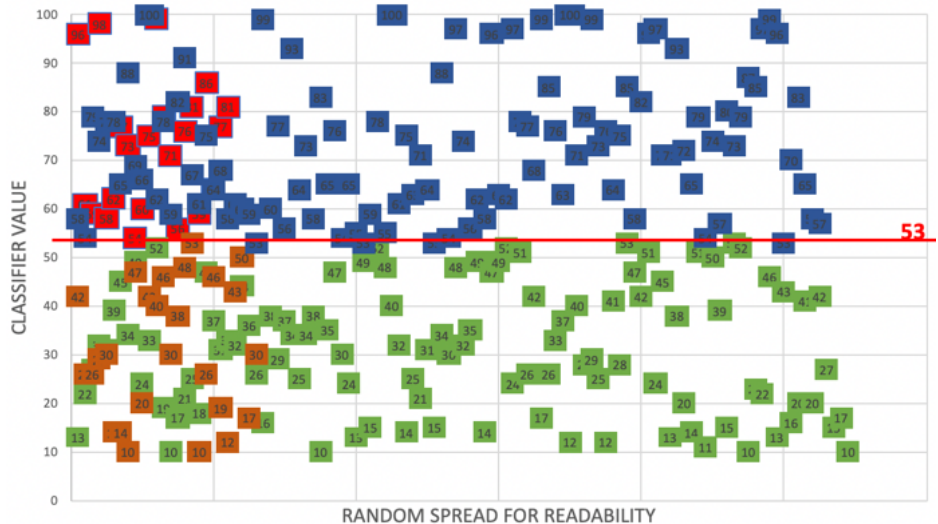

**Figure 1: SVM Results – red unwell misdiagnosed, orange healthy misdiagnosed, green unwell properly diagnosed, blue healthy properly diagnosed.**

This result was slightly improved upon by using the BERT architecture model: 66% were correctly classified, 16% were unwell but misdiagnosed, and the remaining 18% were healthy. This result is presented in more details in Figure 2. The optimal division of space for classification was 60 (note red line).

The use of Collgram analysis also resulted in a slight improvement upon the baseline result, but an important aspect of this improvement is the fact that the number of cases of misdiagnosed patients was noticeably smaller than in the case of the BERT architecture. Specifically, 69% were correctly classified, 12% of texts were generated by depressive subjects but misdiagnosed, and the remaining 19% of entries generated by healthy individuals were misdiagnosed as depressed. The limit value was 79. These results are presented in Figure 3.

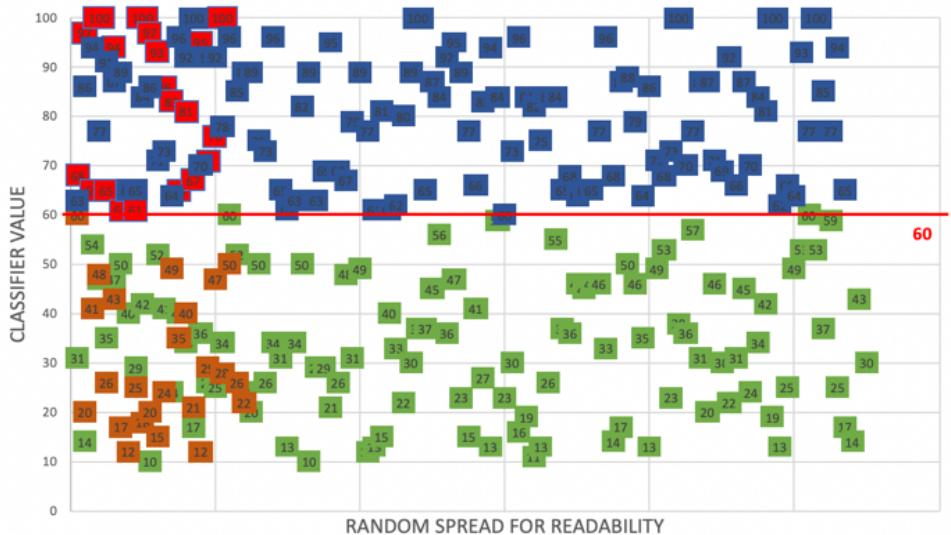

**Figure 2: BERT Results – red unwell misdiagnosed, orange healthy misdiagnosed, green unwell properly diagnosed, blue healthy properly diagnosed.**

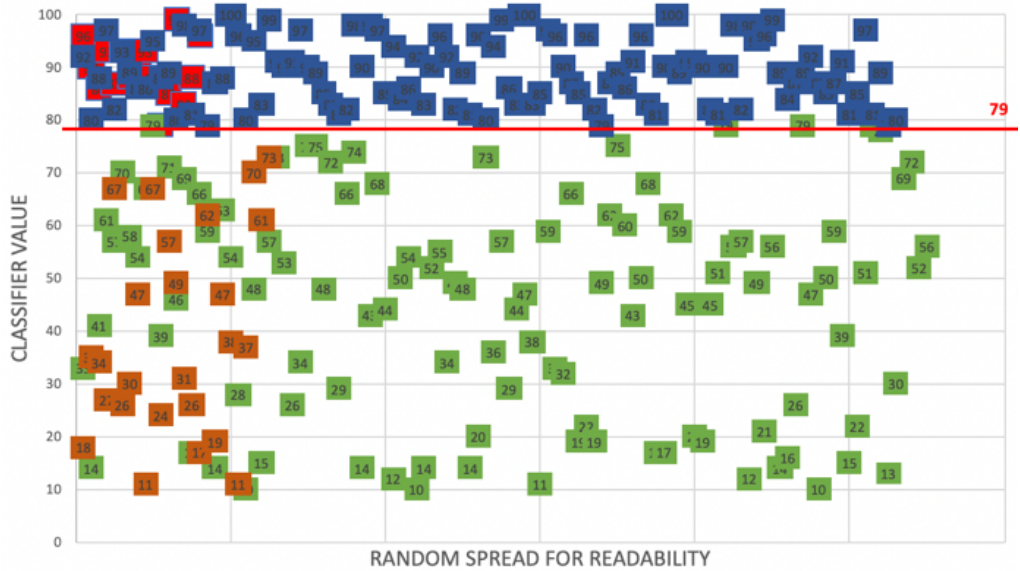

**Figure 3: Collgram Results – red unwell misdiagnosed, orange healthy misdiagnosed, green unwell properly diagnosed, blue healthy properly diagnosed.**

A hybrid approach was also tested, where the results of the BERT and Collgram models were averaged arithmetically. This improved our results further by 2%. More specifically, 71% of cases were correctly classified, with 10% being false negatives, and 19% being false positives. The optimal limit value was 68, which is shown in Figure 4.

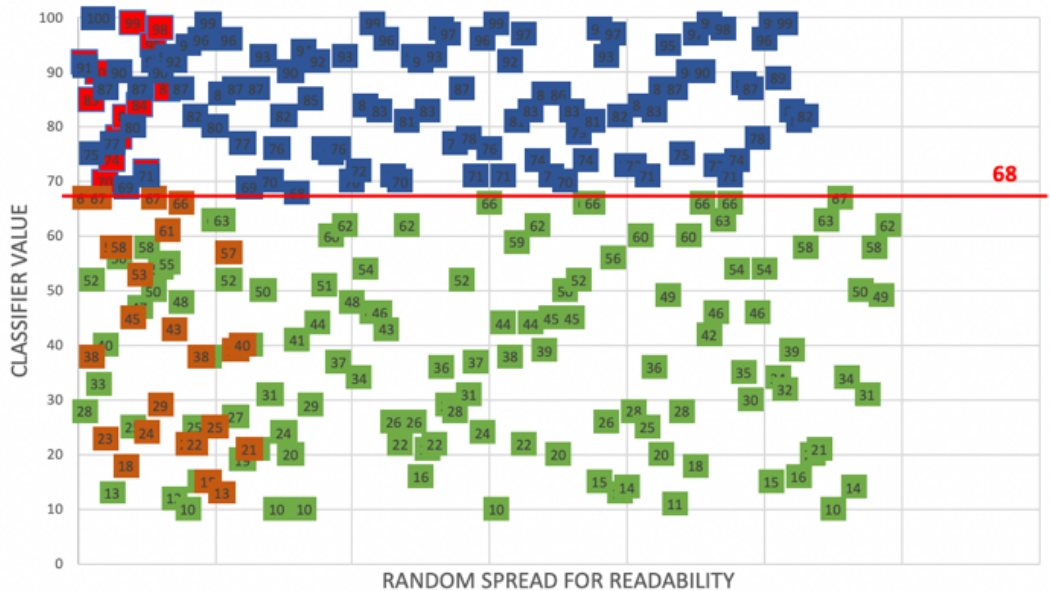

**Figure 4: Hybrid Results – red unwell misdiagnosed, orange healthy misdiagnosed, green unwell properly diagnosed, blue healthy properly diagnosed.**

Results are summarized in Table 2.

|              | **Correctly diagnosed** | **Unwell, misdiagnosed** | **Healthy** |
|--------------|-------------------------|--------------------------|-------------|
| SVM          | 63%                     | 17%                      | 20%         |
| BERT         | 66%                     | 16%                      | 18%         |
| Collgram     | 69%                     | 12%                      | 19%         |
| BERT+Collgram| 71%                     | 10%                      | 19%         |

**Table 2: Summarized results.**

## Discussion and conclusions

In this research paper we applied Collgram analysis to understand how texts were constructed and what their characteristics were, rather than applying a traditional approach to analyse specific words or sentences. Collgram analysis proved to be an effective method to identify symptoms of depression in social media entries. We also found that when conducted together, Collgram and sentiment analyses allow for a better accuracy than each of the methods separately. It suggests that combined approach is more effective in assessment of affective disorders based on NLP text analysis, but more research is needed to strengthen this claim.

Conducting observational and longitudinal studies are the other advantages of the proposed hybrid approach, allowing to assess social media entries of given an individual, or groups of individuals over a longer period. The possible applications include evaluation of the cause of the diseases and the effectiveness of psychotherapy or pharmacotherapy.

The other practical application of our method in clinical practice could be to support diagnostic process in the clinical setting. It could help in objectifying the clinical assessments made by mental health practitioners. Our method could also be used to analyse speech samples, after pre-processing them with an automated speech recognition system, similarly to (Sehgal et al. 2018).

The main limitation of our proposed method is that it is not language independent and it requires a relatively complex pre-processing of the data.

To conclude, we demonstrated a proof-of-concept, and practical applicability of our hybrid method detecting symptoms of depression in social media entries. We believe that this computer-based method can be used in an innovative mental health decision support system.

## Acknowledgements

We thank Michael Connolly for proofreading.

## References

Al Hanai, T., Ghassemi, M. M., Glass, J. R. 2018. "Detecting Depression with Audio/Text Sequence Modeling of Interviews." In Interspeech (Vol. 2522), pp. 1716-1720.

Beck, A. T., Steer, R. A., Beck, J. S., Newman, C. F. 1993. "Hopelessness, depression, suicidal ideation, and clinical diagnosis of depression," Suicide and Life-Threatening Behavior 23, 2 (1993), 139–145.

Bestgen, Y., Granger, S. 2014. "Quantifying the development of phraseological competence in L2 English writing: An automated approach." Journal of Second Language Writing, pp. 26, 28-41.

Bestgen, Y., Granger, S. 2018. "Tracking L2 writers' phraseological development using collgrams: Evidence from a longitudinal EFL corpus." In Corpora and lexis. Brill Rodopi. pp. 277-301

Bevendorff, J., Stein, B., Hagen, M., Potthast, M. 2018. "Elastic chatnoir: Search engine for the clueweb and the common crawl." In European Conference on Information Retrieval. Springer, Cham. pp. 820-824

Daudaravičius, V., Marcinkevičienė, R. 2004. "Gravity counts for the boundaries of collocations." International Journal of Corpus Linguistics, 9(2), 321-348.

De Schuymer, B., De Meyer, H., De Baets, B., Jenei, S. 2003. "On the cycle-transitivity of the dice model." Theory and Decision, 54(3), pp. 261-285.

Devlin, J., Chang, M. W., Lee, K., Toutanova, K. 2019. "BERT: Pre-training of Deep Bidirectional Transformers for Language Understanding." In Proceedings of the 2019 Conference of the North American Chapter of the Association for Computational Linguistics: Human Language Technologies, Volume 1 (Long and Short Papers) pp. 4171-4186.

Holas, P., Krejtz, I., Wisiecka, K., Rusanowska, M., & Nezlek, J. B. (2020). Modification of Attentional Bias to Emotional Faces Following Mindfulness-Based Cognitive Therapy in People with a Current Depression. Mindfulness, 11. 1413–1423.

Islam, M. R., Kabir, M. A., Ahmed, A., Kamal, A. R. M., Wang, H., Ulhaq, A. 2018. "Depression detection from social network data using machine learning techniques." Health information science and systems, 6(1), 8.

Jung, H., Park, H. A., Song, T. M. 2017. "Ontology-based approach to social data sentiment analysis: detection of adolescent depression signals." Journal of medical internet research, 19(7), e259.

Krejtz, I., Holas, P., Rusanowska, M., Nezlek, J. (2018). Positive online attentional training as a means of modifying attentional and interpretational biases among the clinically depressed: An experimental study using eye tracking.. Journal of Clinical Psychology. 2018 Mar 15. DOI: 10.1002/jclp.22617.

Kroenke, K., Strine, T. W., Spitzer, R. L., Williams, J. B., Berry, J. T., Mokdad, A. H. 2009. "The PHQ-8 as a measure of current depression in the general population." Journal of affective disorders, 114(1-3), pp. 163-173.

Lee, J. S., Hsiang, J. 2020. "Patent claim generation by fine-tuning OpenAI GPT-2." World Patent Information, 62, 101983.

Lenko-Szymanska, A., Wolk, A. 2016. „A corpus-based analysis of the development of phraseological competence in EFL learners using the CollGram profile." In the Conference of the Formulaic Language Research Network (FLaRN), Vilnius, pp. 28-30.

MacLeod, C., Mathews, A. 2012. "Cognitive bias modification approaches to anxiety." Annual review of clinical psychology, 8, 189-217. DOI/10.1146/annurev-clinpsy-032511-143052

McGinnis, R. S., McGinnis, E. W., Hruschak, J., Lopez-Duran, N. L., Fitzgerald, K., Rosenblum, K. L., and Muzik, M. 2018. "Rapid anxiety and depression diagnosis in young children enabled by wearable sensors and machine learning," In 2018 40th Annual International Conference of the IEEE Engineering in Medicine and Biology Society (EMBC). IEEE, 3983–3986.

Morales, M. R., Levitan, R. 2016. "Speech vs. text: A comparative analysis of features for depression detection systems." In 2016 IEEE Spoken Language Technology Workshop (SLT). IEEE. pp. 136-143

Peng, Y., Yan, S., Lu, Z. 2019. "Transfer Learning in Biomedical Natural Language Processing: An Evaluation of BERT and ELMo on Ten Benchmarking Datasets." In Proceedings of the 18th BioNLP Workshop and Shared Task, pp. 58-65.

Sehgal, R. R., Agarwal, S., Raj, G. 2018. "Interactive voice response using sentiment analysis in automatic speech recognition systems." In 2018 International Conference on Advances in Computing and Communication Engineering (ICACCE), pp. 213-218.

Song, L., Langfelder, P., Horvath, S. 2012. "Comparison of co-expression measures: mutual information, correlation, and model based indices". BMC bioinformatics, 13(1), 328.

Sun, C., Huang, L., Qiu, X. 2019. "Utilizing BERT for Aspect-Based Sentiment Analysis via Constructing Auxiliary Sentence." In Proceedings of the 2019 Conference of the North American Chapter of the Association for Computational Linguistics: Human Language Technologies, Volume 1 (Long and Short Papers), pp. 380-385.

Wolohan, J. T., Hiraga, M., Mukherjee, A., Sayyed, Z. A., Millard, M. 2018. "Detecting linguistic traces of depression in topic-restricted text: Attending to self-stigmatized depression with NLP." In Proceedings of the First International Workshop on Language Cognition and Computational Models, pp. 11-21.

Wołk, A., Wołk, K., Marasek, K. 2017. „Analysis of complexity between spoken and written language for statistical machine translation in West-Slavic group." In Multimedia and network information systems. Springer, Cham. pp. 251-260

Wołk, K. 2019. "Advanced social media sentiment analysis for short-term cryptocurrency price prediction." Expert Systems, e12493.

World Health Organization. 2008. The Global Burden of Disease 2004 update. Retrieved from https://www.who.int/healthinfo/global_burden_disease/GBD_report_2004update_full.pdf.

Zaśko-Zielińska, M., Piasecki, M. 2015. „Lexical Means in Communicating Emotion in Suicide Notes– on the Basis of the Polish Corpus of Suicide Notes". Cognitive Studies, (15).

Zucco, C., Calabrese, B., Cannataro, M. 2017. "Sentiment analysis and affective computing for depression monitoring." In 2017 IEEE International Conference on Bioinformatics and Biomedicine (BIBM). IEEE. pp. 1988-1995